\documentclass[10pt,twocolumn,letterpaper]{article}

\usepackage{iccv}
\usepackage{times}
\usepackage{epsfig}
\usepackage{graphicx}
\usepackage{amsmath}
\usepackage{amssymb}
\usepackage{booktabs}
\usepackage{multirow}
\usepackage{multirow}
\usepackage{amsfonts}
\usepackage{graphics}
\usepackage{algorithm}
\usepackage{algorithmic}
\usepackage{xcolor}

\usepackage[pagebackref=true,breaklinks=true,letterpaper=true,colorlinks,bookmarks=false]{hyperref}

\iccvfinalcopy 

\def\iccvPaperID{10774} 

\ificcvfinal\fi

\begin{document}

\title{Towards Human-Understandable Visual Explanations: 
\\Imperceptible High-frequency Cues Can Better Be Removed}

\author{Kaili Wang*$\dagger$ \\
\and
Jose Oramas$\dagger$\\
*KU Leuven, ~~~$\dagger$ University of Antwerp, imec-IDLab\\

\and
Tinne Tuytelaars* \\

}

\maketitle
\ificcvfinal\thispagestyle{empty}\fi

\begin{abstract}
Explainable AI (XAI) methods focus on explaining what a neural network has learned - in other words, identifying the features that are the most influential to the prediction. In this paper, we call them "distinguishing features". 
However, whether a human can make sense of the generated explanation also depends on the perceptibility of these features to humans.
To make sure an explanation is human-understandable, we
argue that the capabilities of humans, constrained by the Human Visual System (HVS) and psychophysics, need to be taken into account. 
We propose the {\em human perceptibility principle for XAI}, stating that, to generate  human-understandable  explanations, neural networks should be steered towards  
focusing on human-understandable cues during training.
We conduct a case study 
regarding the classification of real vs. fake face images, where many of the distinguishing features picked up by standard neural networks turn out not to be perceptible to humans. By applying the proposed principle, a neural network with human-understandable explanations is trained which, in a user study, is shown to better align with human intuition. This is likely to make the AI more trustworthy and opens the door to humans learning from machines. 
In the case study, we specifically investigate and analyze the behaviour of the human-imperceptible high spatial frequency features in neural networks and XAI methods.

\end{abstract}

\section{Introduction}
\label{sec:intro}
\begin{figure}
\centering
\includegraphics[width=0.45\textwidth]{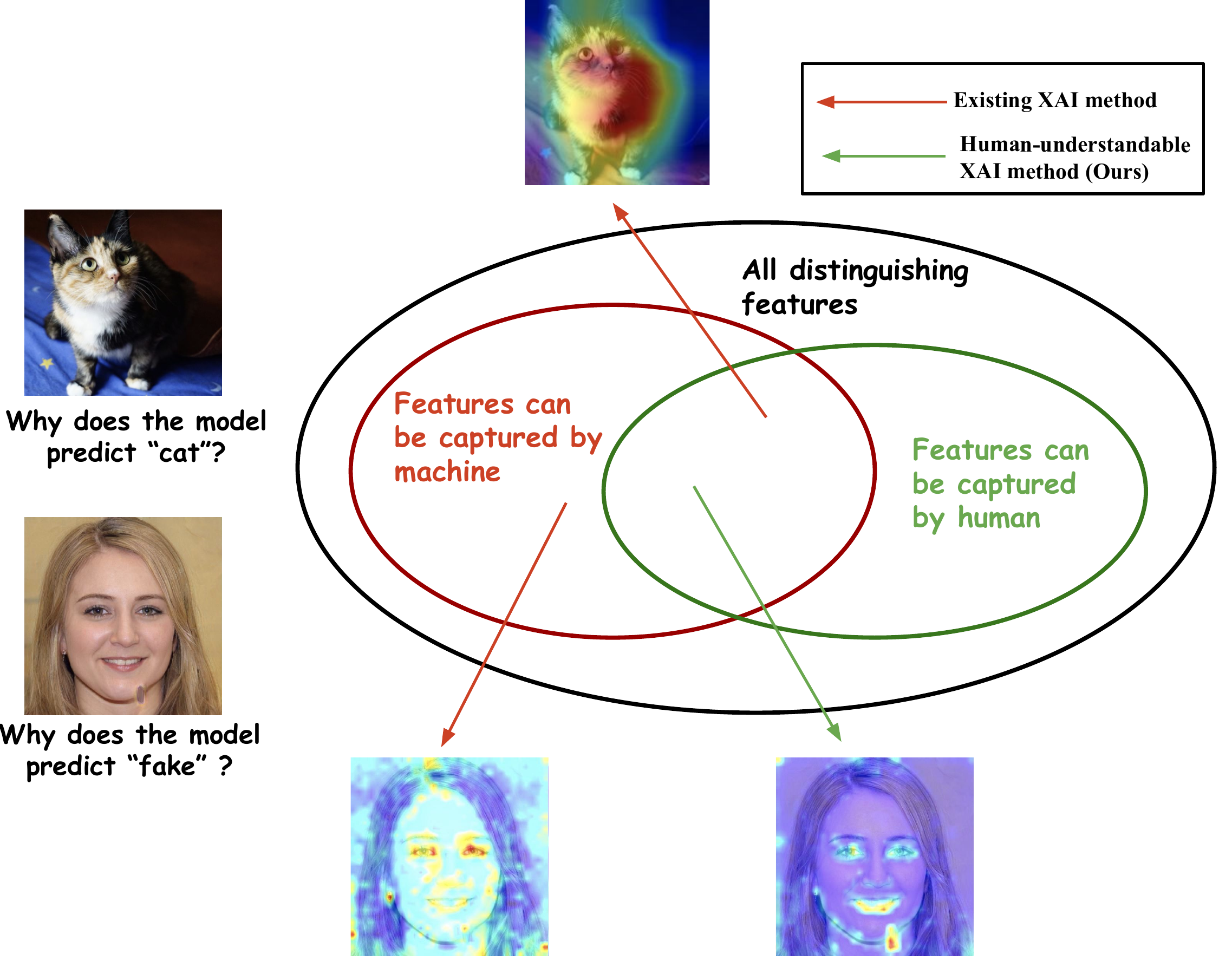}
\caption{
The perception mechanism for human and machine is different.
Existing XAI methods focus on explaining the machine. Humans can only understand their explanation in as far as the model picks up distinguishing features that can be perceived by both machine and human.
However, for some datasets distinguishing features used by the machine are not perceptible to humans.
In this paper we propose a method to generate more human-understandable models and explanations on these datasets.
Please zoom in to check the detail for the fake face example.
}
\label{fig:teaser}
\end{figure}
Most of existing heatmap-based XAI methods such as LRP \cite{lrp_15}, GradCAM \cite{Gradcam17} or, more recently, Vision Transformer (ViT)'s attention \cite{vit_dosovitskiy2020} have shown their ability on neural network explanation for traditional classification tasks, e.g.~on CIFAR \cite{durall2019unmasking}, ImageNet \cite{imagenet_cvpr09} or CUB \cite{WelinderEtal2010}.
The generated explanations from these methods highlight the (most discriminative) region of the predicted class in the image and align well with human intuition.
%
These popular classification datasets have been chosen to evaluate XAI because it is clear, for humans, which areas should be highlighted, providing a mechanism to evaluate the XAI methods.
They have 
a relatively large inter-class visual variance 
(e.g.~cat vs.~car vs.~fish in ImageNet) or, 
for fine-grained datasets like CUB, even though all the images are birds, the visual features defining the classes are known and determined, e.g. color, shape, and texture of particular parts.
We refer to the features that define the difference between classes as \textit{distinguishing features}.
The main characteristic of these popular classification datasets is that 
\textit{the distinguishing features are perceptible by humans.} 
Summarily, humans can understand most of the explanations generated from the existing methods on these datasets because \textit{the distinguishing features are  perceptible by both the current neural networks and humans.}
The cat image in Fig.~\ref{fig:teaser} is an example.


However, 
for other datasets, the visual inter-class variance is small, 
and the distinguishing features may not 
immediately be obvious to
humans.
For instance, consider the distinction between GAN-generated (fake) face images and real face images.
It is under these circumstances, especially, that explanations of automated decisions become important. 
Yet unlike the cat example in Fig. \ref{fig:teaser}, the fake face example shown below leads to an unintuitive explanation: the generated heatmap based on the same method (left) highlights almost the entire image, leaving humans confused about the explanation and leading to reduced rather than increased trust in the AI. This is in contrast to the heatmap generated using our method (right), which is more intuitive and helps humans to understand the decision. 

We therefore introduce the {\bf Human Perceptibility Principle for XAI}:
\textit{To generate human-understandable explanations, a neural network should be steered towards using features that are perceptible by humans.}

To this end, we revisit models of the Human Visual System and pyschophysics, in order to identify the main flaws of human vision.
One of them is the poor perceptibility of high spatial frequency features~\cite{Penn92}.
On the contrary, neural networks, especially convolutional neural networks (CNN), 
have a tendency to focus on high spatial frequency features~\cite{Geirhos19_textureBiased,Wang_2020_CVPR}.
When a major distinguishing feature is related to high spatial frequency, the explanation from current methods can therefore be difficult for humans to understand.

We illustrate these ideas based on a case study: explaining the classification model for fake vs.~real face images.
We believe this is important and instructive since a more human-understandable explanation can help 
non-experts
recognize fake face images in daily life while current explanation methods fail to do so.
To evaluate our method, we conduct a 
user study, asking users' opinion on the generated explanations of 
our method and the method without applying our proposed principle (vanilla method).


In summary, this paper makes the following contributions:
i) 
we focus on making XAI methods more human-understandable and propose the \textit{Human perceptibility principle}, which, we believe, can be an important complement to current XAI research. 
ii) We theoretically and empirically show that human and neural networks have different preferences on capturing features from images, 
especially with respect to the high spatial frequency features, 
and propose a principle to generate human-understandable explanations based on the foundations of the Human Visual System and the psychophysics model.
iii) We study the deepfake case where current XAI methods fail to generate human-understandable explanations. 
The qualitative results and our user study clearly show the explanations can be made more perceptible/understandable for humans by applying the proposed approach.
In addition, 
we investigate and analyze the ability of recent vision transformers at learning/encoding and explaining 
high spatial frequency features.

\section{Related Work}
\label{sec:related}

\subsection{Faithful model explanation}


Input-modification-based XAI methods~\cite{Zeiler_14,Taxonomy_XAI_GrunRNT16, meaningfulPeturbations_FongV17}, a.k.a. occlusion / perturbation-based methods, systematically occlude or modify parts of the input with the goal of modelling how modifications in the input space affect the network predictions.
They operate under the assumption that there is no access to the internal states of the model to be explained, treating it as a black-box. As a consequence, they possess reduced guarantees regarding to the faithfulness of the explanation w.r.t.~the inner-workings of the model.

Based on the previous observation, another group of work \cite{Simonyan14a,smoothGrad_Smilkov17,Zeiler_14,Gradcam17,kaili:visualExpInter} has been proposed with the goal of optimizing the faithfulness of the explanations  w.r.t.~the model being explained. 
By focusing on faithfulness, these methods successfully shed light into some of the internal decision-making processes of the networks. However they suffer from  low-intelligibility and ambiguity in the produced explanations.
All these works fixate on one direction, that is finding and visualizing the distinguishing features taken from the neural network when it is trained for a classification task.   
Different from them, we aim at finding a trade-off where the cues highlighted by the generated explanations are both important for the decision-making process of the network, and understandable by non-expert users. 
Moreover, while these methods operate in a post-hoc manner, the proposed principle aims at the design of models whose representation is interpretable-by-design.

\subsection{Concept-based Explanations}
Following the previous efforts, a new group of methods has emerged which aim at generating explanations based on "concepts" derived from the model being explained.

Towards this goal, \cite{TCAV_KimWGCWVS18} trains linear classifiers to derive concept vectors and link the importance of each concept with the classes of interest.
\cite{kaili:visualExpInter} identifies sparse features encoded in the model that are relevant for the classes of interest. These features are then used as means for explanation.
\cite{ghorbani2019automatic} 
compute superpixels from each image example. Then, concepts are identified by clustering all the superpixels, from the entire dataset, that are important for the model.
The methods listed above use intermediate concepts extracted from internal activations from the model. Therefore, while capable of providing an intuition, they do not necessarily possess a direct semantic meaning, thus, reducing their intelligibility.

To address this issue,
\cite{IBD_ZhouSBT18} proposes decomposing neural activations from images into semantically interpretable components, pre-trained from a large concept corpus. Then, explanations are obtained by projecting the feature vector into the learned interpretable basis.
Very recently, \cite{ConceptBottleneckModels_KohNTMPKL20} has revisited the idea of predicting concepts, provided at training time, and then uses these concepts to predict the label. 
These works enable attaching understandable text-based concepts as part of the output of a model. On the downside, this capability comes at the cost of additional semantic concept annotations for training. Moreover, for \cite{IBD_ZhouSBT18} expensive pixel-level annotations are also required. 

Different from them, we investigate an orthogonal direction where we stress that explanation of distinguishing features must first be perceptible to humans in order to grant the characteristic of being intelligible.

\section{Methodology}
\label{sec:method}



\subsection{Problem Statement}
\label{sec:problemstatement}
\begin{figure}
\centering
\includegraphics[width=0.48\textwidth]{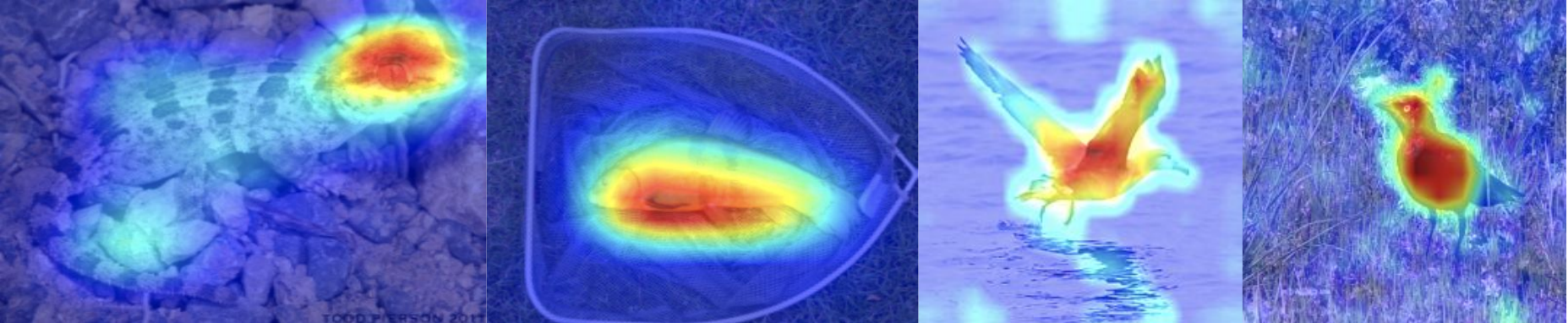}
\caption{
CAM-based method and ViT Attention-based method applied on the Imagenet dataset (left two) and CUB dataset (right two). People can easily understand the relationship between the heatmap and its corresponding class.  
}
\label{fig:cub_example}
\end{figure}

\begin{figure*}
\centering
\includegraphics[width=1\textwidth]{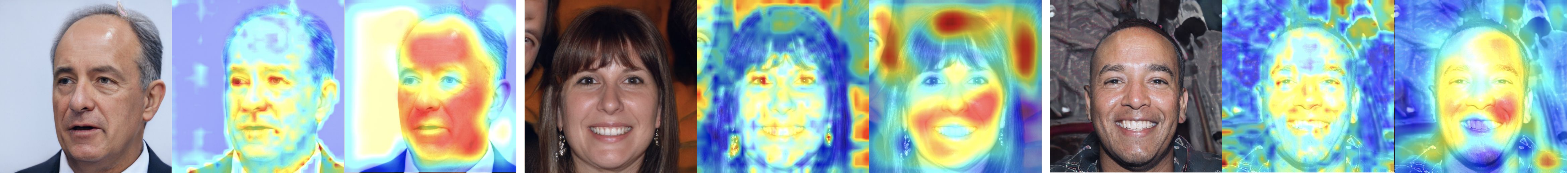}
\caption{ViT attention (middle) and CAM visualization (right) for three styleGAN-generated fake images, based on the neural networks that classify the fake and real images.
Can you understand which features from the input fake images the highlighted regions indicate?
}
\label{fig:failure_example}
\end{figure*}

Existing heatmap-based model explanation methods try to explain the decision made by a  pretrained neural network. In other words, the goal is to identify the most important features (for the network's task) extracted by the neural network from the input image.
Fig.~\ref{fig:cub_example} shows some examples of GradCAM \cite{Gradcam17} applied on the Imagenet dataset.
People can easily understand the explanation since the distinguishing features are both perceptible by the machine and humans. 
Fig.~\ref{fig:failure_example} shows some examples of GradCAM and ViT \cite{vit_dosovitskiy2020,abnar2020quantifying} attention applied on a deepfake dataset \cite{styleGAN2018}, where the model is trained to classify images as fake or real.
Obviously, these methods fail. Not only do they generate very different explanations focusing on different image regions / distinguishing features, it is also difficult for humans to understand the generated explanations - or tell which one to trust more. 
Therefore, the main research question of this paper is: \textit{Is there a method that can generate human-understandable explanations for this type of data?}

\subsection{Revisiting Human Visual System and Psychophysics Model}
To answer the question, we need to identify the features that can / cannot be perceived by humans. 
Therefore, we revisit the Human Visual System (HVS) and Psychophysics model and discuss the main characteristics related to human perception on digital images. We limit ourselves to the essence and do not elaborate on the biology principles behind them. 

\textbf{Luminance and Color}
The HVS has more resolution on luminance than chroma information. This is one of the motivations of chroma subsampling in image compression field \cite{Winkler01}.
In addition, due to the features of the cone, the HVS can perceive a limited variety of colors, estimated around 10 million.
On the contrary, we know color can be represented as a numerical combination (e.g. RGB color space) in a machine, which leads to more than 16 million colors (with 8 bit per channel). The total number is clear much larger than what can be perceived by humans.

\textbf{Weber Law}
Weber Law describes that the Just-Noticeable Difference (JND) $dS$ is proportional to the initial stimuli intensity $S$, $dS = K S$, where $K$ is a constant. JND is the smallest change in stimuli that can be perceived by humans \cite{Kandel13}.
%
In the context of digital images, Weber Law explains why humans are more sensitive to detailed regions, e.g. texture, edges, where there is a relative large difference around the neighbouring pixels, rather than flat regions.
Likewise, structures in dark image areas are often missed.

\textbf{Perceived Spatial Frequency}
Weber Law indicates that humans are more sensible on detailed regions of an image. However, subtle differences in regions with very high spatial frequency are not perceptible by humans. 
For instance, JPEG~\cite{Penn92} takes advantage of this characteristic to quantize these high frequency components without a perceptible loss of quality by humans.
In short, high frequency components are less perceptible to a human.

\subsection{Revisiting Machine Vision}
The most significant difference w.r.t.~the HVS is that images are stored and processed as numbers in a machine.
In other words, the perceived mechanism is totally different from the HVS.
Therefore, machines can easily distinguish very similar colors simply because the numbers of their RGB representation are different.

Here we focus the discussion on convolutional neural networks (CNN).
We treat transformers  
as one kind of CNN since the convolutional blocks are still used in the architecture. 
Please refer to the original paper~\cite{vit_dosovitskiy2020} for more technical details.
Different from the HVS, CNNs have a good ability to capture high spatial frequency.
\cite{Geirhos19_textureBiased,liu2020global,Wang_2020_CVPR} have shown that CNNs leverage texture information significantly for classification tasks to the point of even being biased towards texture rather than shape \cite{Geirhos19_textureBiased}.
\cite{kaili:visualExpInter} also shows that front layers of CNNs usually capture low-level features, e.g. color and texture.

\subsection{Proposed Approach}
In this section,
we first describe the Human Perceptibility Principle for XAI in more detail.
%
%
Given a dataset $D$ and corresponding classification task $T$, there exists a set of distinguishing features $\Phi = [ \phi_1, \phi_2, ..., \phi_n]$ based on which the classes in $D$ can be distinguished (i.e.~the classification task is feasible). 
In practice, we have a subset  $\Phi^h \subseteq \Phi$ of features perceptible by humans and another subset $\Phi^m \subseteq \Phi$ of features perceptibly by machines.
When training a neural network $f$ on the dataset, it will use a subset $\Psi^m$ of the distinguishing features in $\Phi^m$ to base its decision on. 
If, however, $\Psi^m \cap \Phi^h = \emptyset$, the used features are not perceptible by humans and no human-understandable explanation can be generated. 
%
Therefore, the machine should be steered away from using features that are exclusively perceptible by machine $\Phi^m \setminus \Phi^h$, and stimulated to consider more human-understandable ones $\Phi^m \cap \Phi^h$. 


To achieve this,
there are two possible approaches: i) use a pre-processing technique to get rid of these imperceptible features at the image level (input), ii) use data-augmentation techniques to learn the network to become invariant to differences imperceptible by humans, or iii) during the training process, make the neural network focus less on these features, e.g.~using additional loss terms penalizing the use of such features.
Obviously, this is only possible if indeed, there exists human-understandable distinguishing features $\Phi^h$ for task $T$ in $D$.

Based on the HVS and the psychophysics model, we identify the following characteristics that XAI methods should take into account in order to generate human-understandable explanations:
i) Human eyes cannot distinguish very fine-grained color differences.
ii) Human eyes can hardly perceive the difference in the high spatial frequency components.
iii) Human eyes are sensitive to edge regions.

%
In our study, we focus on high spatial frequency components as the imperceptible distinguishing feature.
We select the first approach and use a \textit{bilateral filter}~\cite{bilateral_98} to process the input images.
It can smooth high frequency regions, narrowing the difference between the frequency distribution of fake and real images, as well as preserve the edges, where humans are sensitive to perceive, which exactly meets the requirement of the HVS.

For popular classification datasets, the main difference between the classes usually lie on the shape, color and/or visible texture. 
This information is perceptible by both machines and humans, and we believe this is the reason why existing XAI methods can generate reasonable good explanations (to humans) on these dataset. 


\section{Experiments}

\subsection{Datasets}

\textbf{FFHQ-HF-WS}
We start with a controlled experiment. To this end, we construct a two-class artificial dataset FFHQ-HF-WS, 
where we mimic two distinguishing features in one class:
high and low spatial frequency features on the images.

\begin{figure}
\centering
\includegraphics[width=0.5\textwidth]{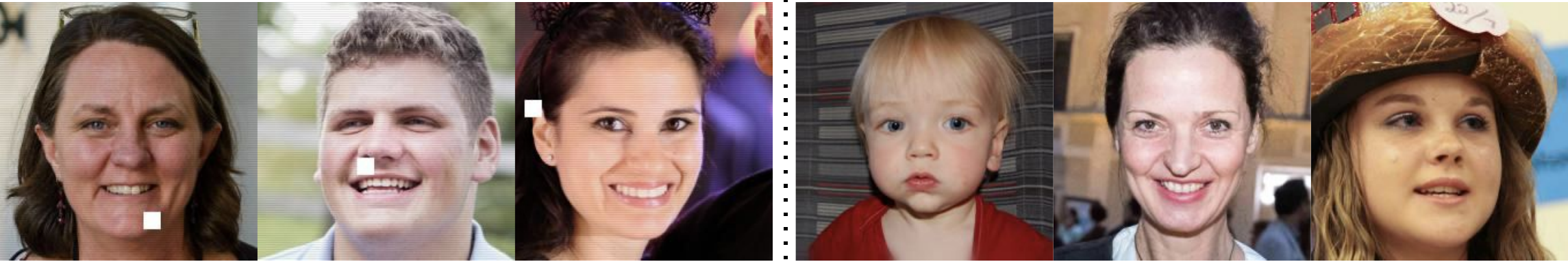}
\caption{Examples for class-1 (left) and class-2 (right) from the FFHQ-HSF-WS dataset. 
}
\label{fig:example-dataset}
\end{figure}

We sample 10K images from the FFHQ dataset \cite{styleGAN2018} and resize them to $224 \times 224$.
For class 1, we add a high spatial frequency feature by periodically changing the intensity of the pixels row by row. More specifically, 
the RGB pixel value is set to 0.9 times  the original value every two rows.
We also add a low spatial frequency cue by putting a $15 \times 15$ white square on the image at a random location. Please note, there is no spatial frequency change inside of the white square.
Some examples of this dataset are illustrated in Fig.~\ref{fig:example-dataset}.
For class 2, we do not introduce any change.
In total, there are 20K images, we use 16K for training and 4K for testing.

\textbf{DeepFakes}
For the realistic GAN-generated deepfake images, we use a subset of Faces-HQ dataset~\cite{durall2019unmasking}, which contains 10K $1024 {\times} 1024$ images from FFHQ dataset as real ones and the same number, and resolution, of images from www.thispersondoesnotexist.com (TPDE) as fake ones.
The fake images are generated by styleGAN~\cite{styleGAN2018}.
In addition, we also use 10K CelebA-HQ~\cite{CelebAMask-HQ} images and 10K fake images generated by styleGAN trained on the CelebA-HQ dataset.
We split them into training, validation and test splits with the proportion of 0.7 : 0.15 : 0.15, respectively.


\subsection{Controlled Experiment}
In this section, we design a controlled experiment to show that 
by removing the human imperceptible  features, machines can shift towards using human-understandable cues (if they exist).

\begin{figure}
\centering
\includegraphics[width=0.45\textwidth]{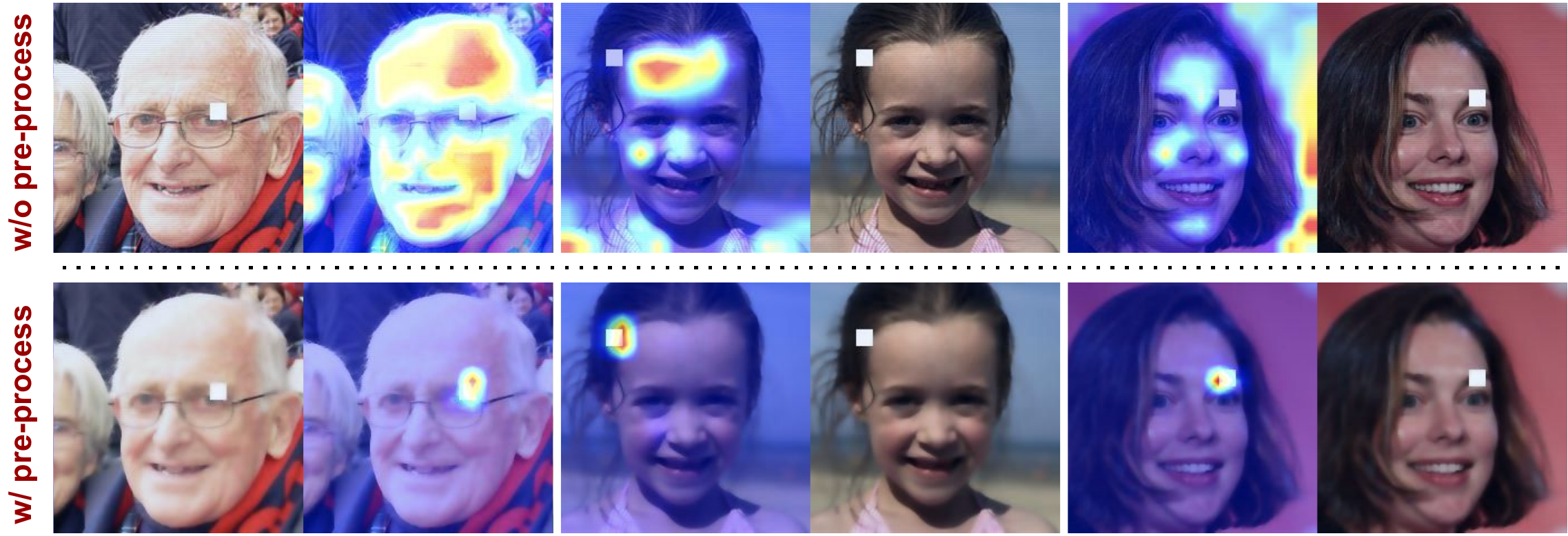}
\caption{
Attention heatmap visualizations. 
The visualizations on top are computed by using the original images for training while for the ones below we apply the bilateral filter as pre-processing.
}
\label{fig:toy_exp}
\end{figure}

We first use our FFHQ-HF-WS dataset to train a binary classifier based on a two-layer ViT-16~\cite{vit_dosovitskiy2020}.
According to the HVS and machine vision, white squares should be picked up by humans more easily while the neural network should prefer the high spatial frequency features.
The classification accuracy is 1 since it is an easy task.
We use the attention as the explanation~\cite{Wiegreffe19,chefer2020transformerInterpretability}. To visualize it we follow \cite{abnar2020quantifying} to roll out the attention. We show some visualizations on the upper part of Fig.~\ref{fig:toy_exp}.
It shows that the heatmaps focus mostly (albeit not exclusively) on some large regions, i.e. face, background, where the high spatial frequency feature is applied and clear, rather than the white square region.
This proves that indeed for CNNs, (ViT here), high spatial frequency features are easier to capture.

Then, in order to
make the neural network take the cue that is 
preferably captured by humans,
 we apply the bilateral filter on the images as a preprocessing step.
After processing the data, we conduct a similar experiment.
The classification accuracy is 99.8\%, slightly lower than the previous one.
Similarly, we visualize the attentions learned by the neural network in the lower part of Fig.~\ref{fig:toy_exp}. Now we can see the attention successfully shifts from the previous high spatial frequency region to the white square region, which aligns better with human perception and, consequently, seems more intuitive to humans.

To quantitatively evaluate the attention heatmap, we calculate its intersection over union (IoU) with the corresponding ground truth (GT) mask.
For high spatial frequency feature, the GT mask is the whole image except the white square while the white square region is the GT mask for the low spatial frequency feature.
Inspired by \cite{choe2020wsol, kaili:visualExpInter}, we use 100 thresholds between 0 and 1 to binarize the generated heatmap and calculate the area under the curve (AUC) of the IoU curve.
Results presented in Table~\ref{tab:auciou} indicate that indeed the attention heatmap shifts from the high spatial frequency region to the white square region which is more human-understandable, after applying the bilateral filter on images. 

\begin{table}
\setlength{\tabcolsep}{4.7pt} 
\scalebox{1}{%
\centering
\begin{tabular*}{8cm}
{@{\extracolsep{\fill}} l c c}
\toprule 
\textit{Experiment}  & \textit{AUC-IoU HSF} & \textit{AUC-IoU WS}   \\ 

\midrule[0.6pt]	
w/o~pre-process filter & 0.18  & 0.00 \\
with pre-process filter& 0.01 & 0.27 \\
\bottomrule[1pt]
\end{tabular*}
}
\vspace{+2mm}
\caption{AUC-IoU result for the two experiments. AUC-IoU HSF indicates the generated heatmap with high spatial frequency mask while AUC-IoU WS refers to the one with the white square mask. }
\label{tab:auciou}
\end{table}

\subsection{Real Case Study: Deepfakes}
In this section, we focus on the deepfake images. More specifically, the images generated by styleGAN~\cite{styleGAN2018}.
Fig.~\ref{fig:failure_example} has shown that current model explanation methods fail to provide human-understandable explanations.
We try to analyze it and generate more human-understandable explanations for these images. We consider this important and instructive, 
since a more human-understandable explanation can help non-experts have a better ability at recognizing these fake face images in daily life.
%
For the sake of simplicity, we mainly use FFHQ/TPDE images for most of our experiments, except for the qualitative results.
Since the results are very similar, we report the experiment based on the CelebA-HQ in the supplementary material.

\begin{figure*}
\centering
\includegraphics[width=0.92\textwidth]{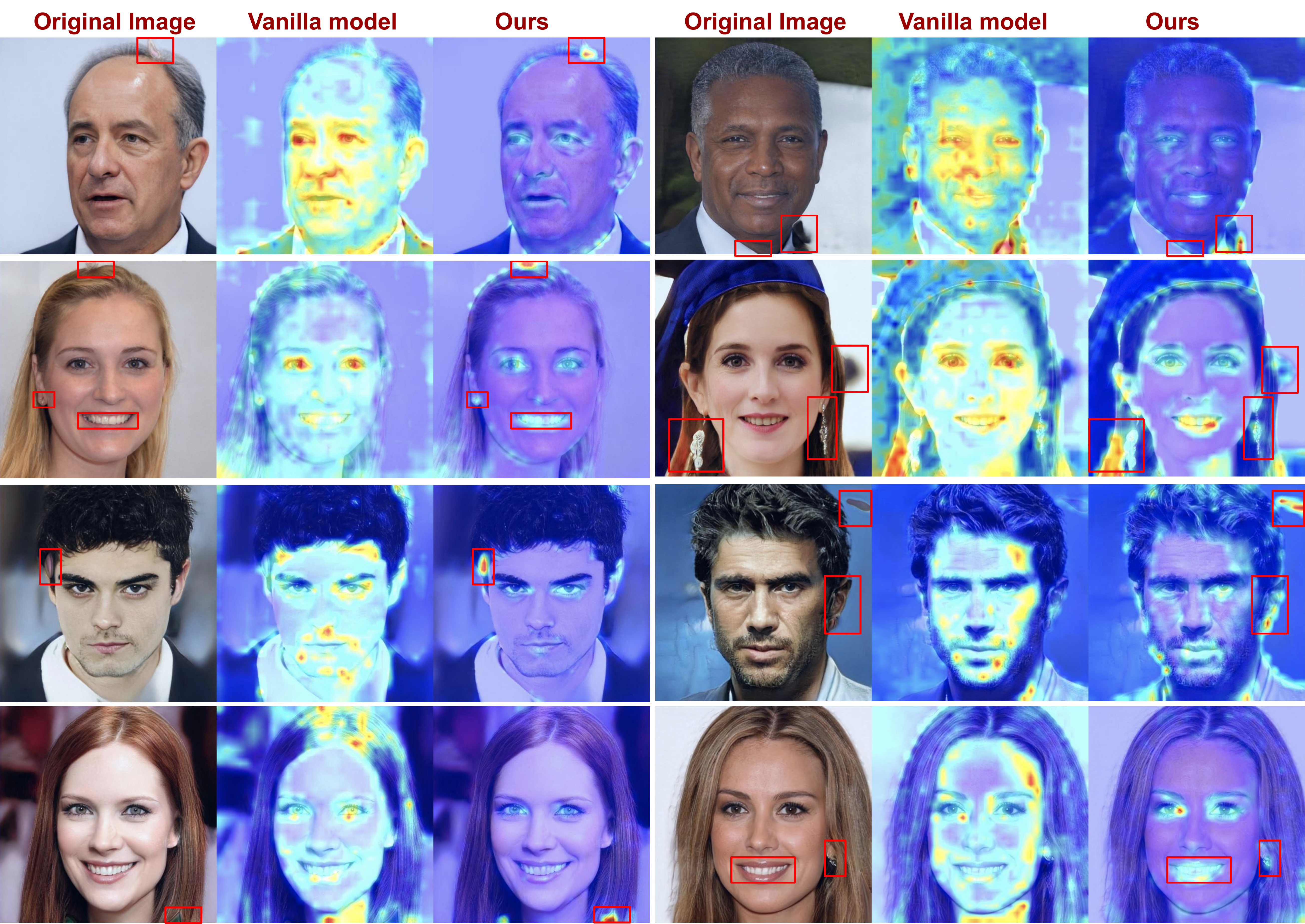}
\caption{
Qualitative comparison for the explanation of \textit{fake} images generated by ViT-16 attention. 
Examples from the TPDE dataset (row 1-2) and styleGAN  trained by celebA-HQ dataset (row 3-4). 
%
Do you feel it makes more sense by looking at the right explanation? 
}
\label{fig:quality_fake}
\end{figure*}
\begin{figure}
\centering
\includegraphics[width=0.36\textwidth]{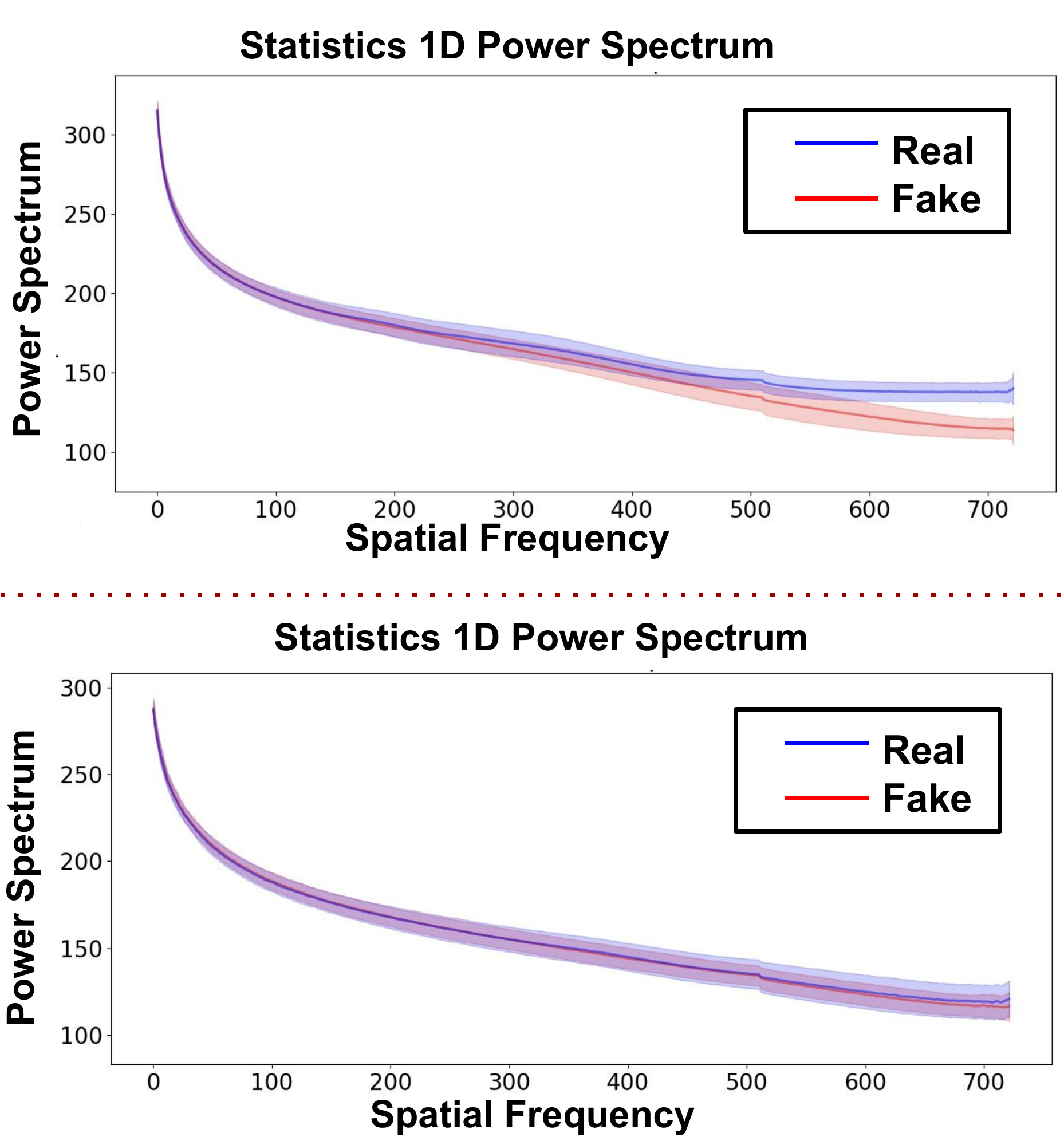}
\caption{
Spatial frequency distributions for real and fake images. Distribution from original images (top) and from images after the pre-processing (down).
}
\label{fig:frequency_dis}
\end{figure}

\textbf{What are the unique features in fake images?}
Recently \cite{durall2019unmasking,durall2020upconv,liu2020global,pmlr-v119-frank20a} showed that the spatial frequency component can be used to distinguish GAN-generated images from real training images, especially the high spatial frequency part.
\cite{durall2019unmasking} even uses a simple support vector machine to successfully classify the real/fake images by extracting and using their frequency components.
This can explain what happens in Fig.~\ref{fig:failure_example}: the heatmap  might  actually be covering the high frequency regions.
However, it is difficult for humans to perceive the high frequency differences.
Similar to \cite{durall2019unmasking}, we analyze the spatial frequency distributions of the real and fake images.
For each image, we take Discrete Fourier Transform to get the amplitude spectrum, then the azimuthal average is applied on the 2D amplitude spectrum and the final 1D spatial frequency distribution is obtained.
Fig.~\ref{fig:frequency_dis} (top) shows the distribution of 100 random images from the real and fake classes.
It is clear that the fake and real images can be distinguished by observing the high spatial frequency part. 


For the human factor, 
\cite{liu2020global} did a user study, where they asked participants their criteria for fake images. The result shows that users normally take cues like “asymmetrical eyes”, “irregular teeth” etc., i.e. the shape and color artifacts, rather than very high spatial frequency pattern.
This is in line with what we concluded from HVS.

\textbf{Generating Human-understandable explanation}
According to our approach, we need to reduce the effect of the high frequency feature on the neural network when it is trained. 
Here we simply use a bilateral filter to pre-process the data.
Fig.~\ref{fig:frequency_dis}(bottom) shows the spatial frequency distribution of these pre-processed images.
Please note that the difference in the high frequency part is significantly reduced.
\footnote{The number of bins in the figures are different since we resize the images to $512 {\times} 512$ in order to fit them into the GPU.}
Here we use the ViT-16 model with 12 layers to train the classifier and use the attention as the explanation.
In addition, in order to accelerate the training process, the model we use is pre-trained on the Imagenet dataset~\cite{vit_dosovitskiy2020}.
The reason we choose Vision transformer is based on the fact that it is patch-based and the human-understandable cues are more local.
We also train a classifier using the unprocessed images as a reference model, we refer to it as the vanilla method.

\begin{figure*}
\centering
\includegraphics[width=0.92\textwidth]{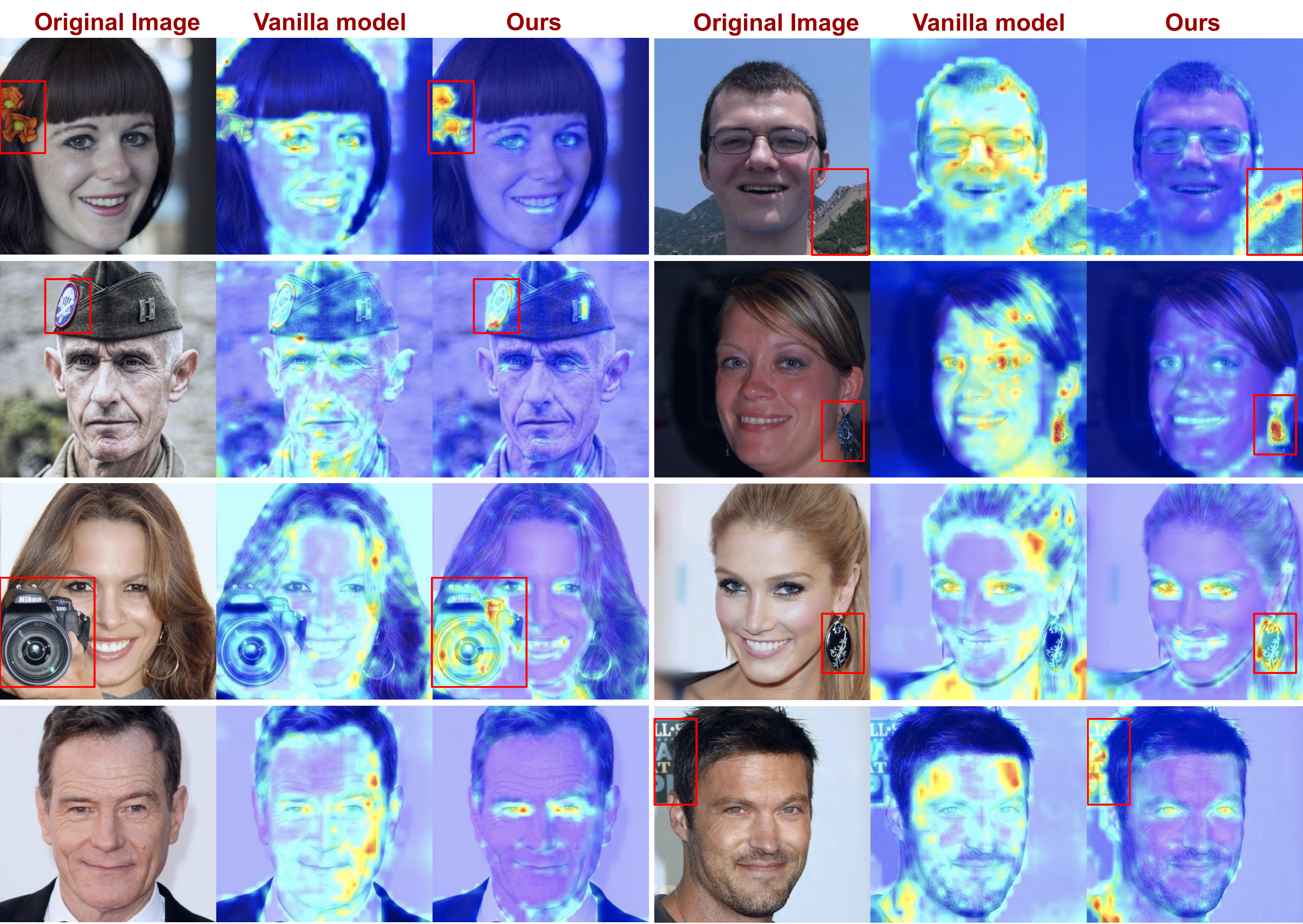}
\caption{
Qualitative comparison for the explanation of \textit{real} images generated by ViT-16 attention. 
The first two rows are from FFHQ while the last two rows are from celebA-HQ dataset.
Do you feel it makes more sense by looking at the right explanation? 
}
\label{fig:quality_real}
\end{figure*}

Fig.~\ref{fig:quality_fake} and Fig.~\ref{fig:quality_real} show several explanation examples for fake and real face images respectively. We use red bounding boxes to indicate some human-understandable regions on these images.
Compared with the explanations from the vanilla approach, our method can indeed localize these human-understandable cues, such as asymmetrical earrings, weird teeth and eyes, and the bubble artifacts for fake images, versus accessories and particular background features for real images.

Then, we calculate the prediction accuracy on the testing split. Table~\ref{tab:prediction_accuracy} indicates that the vanilla method achieves slightly better performance (1.4 pp), which means that for this type of dataset, high spatial frequency feature is a useful cue for the neural network. 
This observation is also in line with the papers \cite{Geirhos19_textureBiased,liu2020global,Wang_2020_CVPR} discussed before.
In addition, to verify the dependency of the vanilla model on high frequency features, 
we measure the performance of the vanilla model when classifying filtered images.
In this setting, the prediction accuracy drops to 50\%, which confirms that indeed the vanilla model takes the spatial frequency feature as the most important cue - without it the model reaches chance levels.

\begin{table}
\setlength{\tabcolsep}{4.7pt} 
\centering
\scalebox{1}{%
\begin{tabular*}{8cm}
{@{\extracolsep{\fill}} l c}
\toprule 
\textit{Experiment}  & Accuracy (\%)\\ 

\midrule[0.6pt]	
Vanilla & 99.97  \\
Ours&   98.57\\
\midrule[0.6pt]	
Vanilla model with filtered images & 50.00\\
\bottomrule[1pt]
\end{tabular*}
}
\vspace{2mm}
\caption{Fake vs real face image prediction accuracy of vanilla model and our model.}
\label{tab:prediction_accuracy}
\end{table}

\textbf{User study}
In order to get human's evaluation on the explanation heatmap we build a website and conduct a user study.
For each fake image, we show the original image and two explanations: one generated by the vanilla model and one based on our model.
The main question asked in the survey is "\textit{Which explanation heatmap is  closer to your idea that the image is fake?}"
We use 100 fake images and its corresponding two explanations, from where we randomly pick 20 images to show to the users each time.
After finishing the questionnaire, we also ask users two questions for feedback: \textit{Q1: Do you feel you got better at recognizing deepfakes based on the test?} and \textit{Q2 Do you feel the heatmaps helped in this process?}

In total, 
84 participants joined the study, 69 of them (82.1\%) think our explanation is closer to what they think (our method has more votes in each survey), 13 of them (15.5\%) prefer the explanation from the vanilla method while there are 2 (2.4\%) users that consider both explanations equally good (The same number of votes for both methods).
On average, our explanations received 71.0\% votes while 29.0\% votes went to the vanilla model.
Regarding the last two questions, 73.2\% of the participants feel they are better at recognizing the fake images after the test, 91.5\% of them think the heatmaps they chose were helpful. 
%
We list the statistics from the study in Table~\ref{tab:user_study}. 
The user study suggests that indeed by removing the imperceptible high spatial frequency feature, the explanations become more human-understandable.
Also, it is interesting to see, that most of the participants feel that they are better at recognizing fake images after doing the test.

\begin{table}
\setlength{\tabcolsep}{4.7pt} 
\centering
\begin{tabular*}{8cm}
{@{\extracolsep{\fill}} l c c}
\toprule 
\textit{Explanation}  & Participants (\%) & Votes (\%)\\ 

\midrule[0.6pt]	
Vanilla & 15.5  &  29.0 \\
Ours &  \textbf{82.1}  & \textbf{71.0}\\
\end{tabular*}
\begin{tabular*}{8cm}
{@{\extracolsep{\fill}} l c}
\toprule 
\textit{Feedback Questions}&Yes (\%)  \\
\midrule[0.6pt]	
\textit{Q1: Better at recognizing fake images?} & 73.2 \\
\textit{Q2: Explanation helpful?} &91.5 \\
\bottomrule[1pt]
\end{tabular*}
\vspace{2mm}
\caption{User study statistics.}
\label{tab:user_study}
\end{table}




\begin{figure}
\centering
\includegraphics[width=0.42\textwidth]{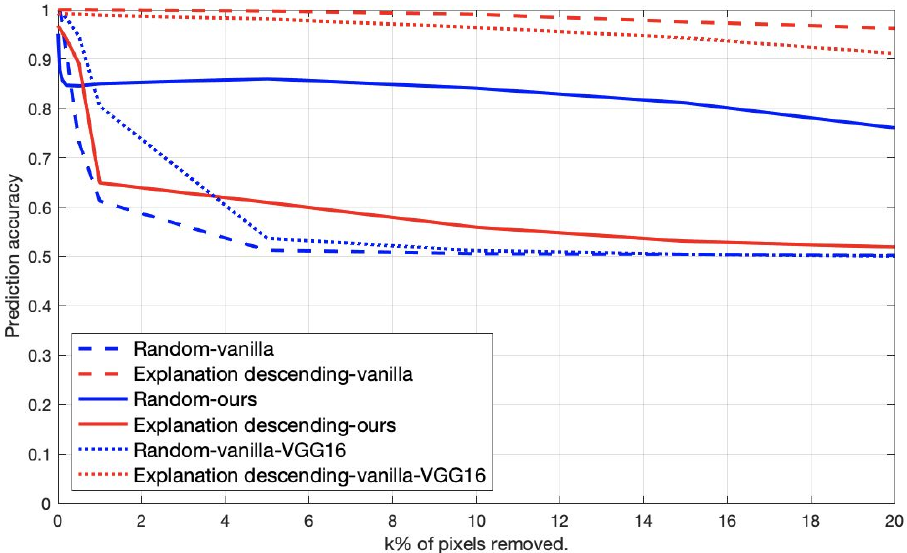}
\caption{
Perturbation test for the vanilla models and our model. 
}
\label{fig:perturbation_curve}
\end{figure}


\section{Validating the Generated Explanations}

In this section, we conduct a perturbation test to study the explanation heatmap and its corresponding model.
For each image $I$, we obtain the attention heatmap $A$ via the model $f$: $A {=} f(I)$.
Then we sort the pixel-level weights in $A$ in descending order.
We gradually remove the top-$k$ important pixels of $I$ according to the relevance-rank indicated in $A$ and send the perturbed input image $I_{p}$ to $f$, obtaining a new prediction.
For reference, we also randomly remove the same amount of pixels on $I$, noted as $I_{p}^{r}$.

Fig~\ref{fig:perturbation_curve} shows the perturbation test result for the vanilla method (dashed line) and our human-understandable method (solid line).
For our method, the trend is clear, the performance of the model decreases significantly as more important pixels are removed. The performance drops to nearly the random guess when the top-20\% of the important pixels are removed while the number is around 75\% for random removal.

It is interesting to see that in the very beginning, the random removal influences the model (to be explained) more than our explanation method.
We think it is because of the architecture of the ViT model whose input is based on several $16 {\times} 16$ patches.
In the very beginning
the random-picked pixels are distributed more separated, which means it can influence more patches at the same time,
while the explanation generated by our method is more localized,
i.e. less patches are influenced.
When $k$ reaches 0.5\%, the performance of our model drops dramatically while the vanilla model keeps stable. 
This experiment proves that the attention heatmap of ViT can be regarded as explanation.

For the vanilla model, surprisingly, we observe an opposite trend.
When $k {=} 20\%$, the random removal experiment has reached the random guess performance while the performance is still quite high (around 99\%) when top 20\% of the most important pixels are removed.
Does it indicate the explanation generated from the vanilla model is useless? How should we interpret it?

\begin{figure}
\centering
\includegraphics[width=0.4\textwidth]{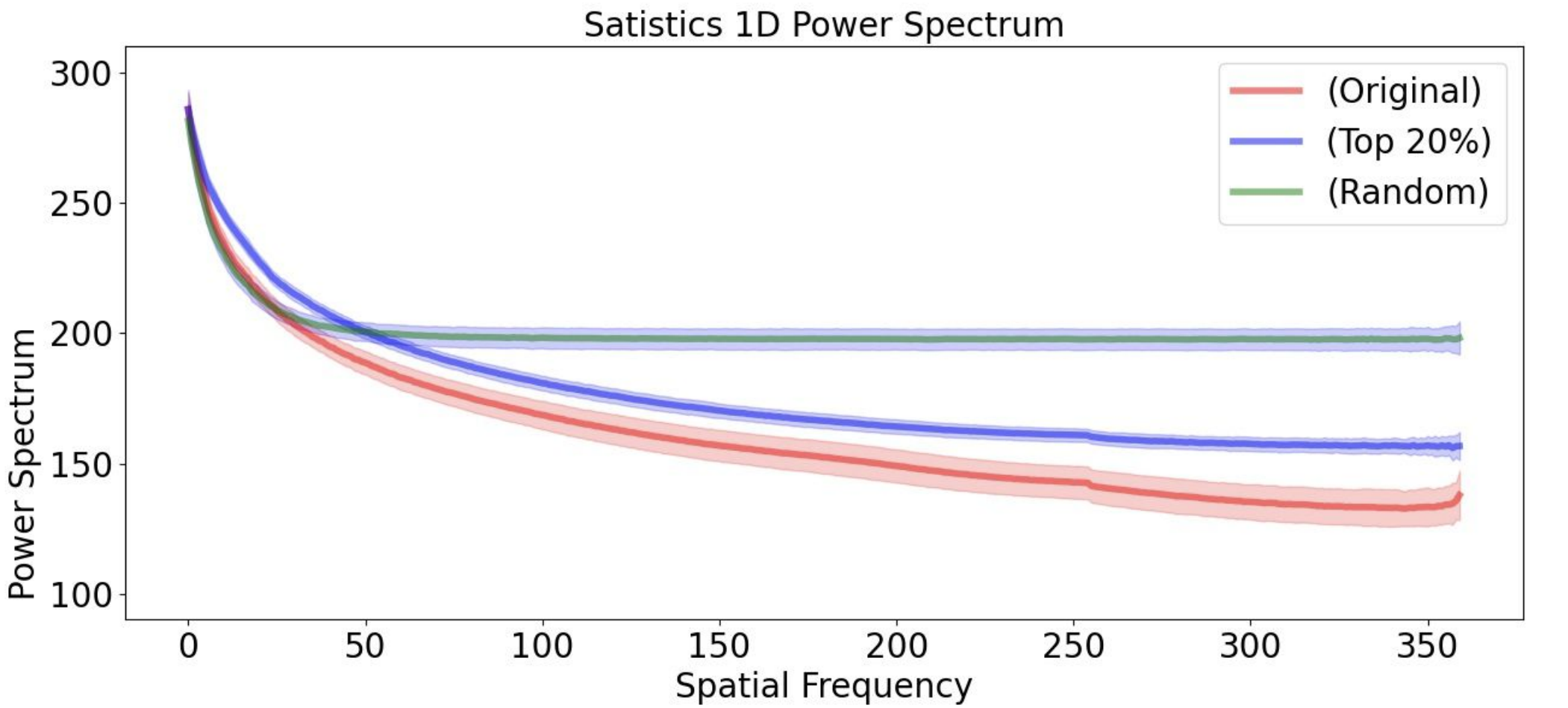}
\caption{
Spatial frequency distribution for $I$ (red), $I_p$ (blue) and $I_{p}^{r}$ (green). 
}
\label{fig:vanilla_remove}
\end{figure}

To answer this question we go back to the spatial frequency of the images.
We empirically prove that the vanilla model mainly takes the spatial frequency feature to do classification.
The spatial frequency feature here is more global, \ie occurs over the whole image.  
Randomly removing pixels can destroy the spatial frequency distribution since these pixels are more separated around the whole image.
On the contrary, the affected pixels are more assembled when top $k$ of the most important ones are removed, \ie less patches are influenced and the spatial frequency distribution does not change significantly. 
Fig.~\ref{fig:vanilla_remove} shows the spatial frequency distribution of the two cases as well as the original images, which implies our analysis is correct. 
In addition, we train a traditional CNN VGG16 in the vanilla manner and obtain a GradCAM explanation.
We repeat the perturbation test on it and observe a similar trend (see the dotted line in Fig.~\ref{fig:perturbation_curve}).  

Therefore, 
the explanation generated for the vanilla method 
can only have the concept level meaning for the global high spatial frequency feature.
The traditional perturbation test that removing pixels according to the weight of explanation map cannot be applied here.

\section{Conclusion}

We propose the \textit{Human Perceptibility Principle for XAI}.
This principle aims at the generation of human understandable explanations by steering the network to use features that 
are perceptible by humans.
%
%
Results from our evaluation suggest that model explanations are indeed more human-understandable when the proposed principle is applied.
In addition, our user study shows that the generated explanations effectively help the participants at understanding the model predictions and can effectively serve as a guide on how to address the task on their own.

{\small
\bibliographystyle{ieee_fullname}
\bibliography{egbib}
}

\end{document}